\documentclass[journal,twoside,web]{ieeecolor_arxiv}

\usepackage{setup/generic}
\usepackage[colorinlistoftodos]{todonotes}

\newacronym{ai}{AI}{Artificial Intelligence}
\newacronym{xai}{XAI}{Explainable \glsentrylong{ai}}
\newacronym{rl}{RL}{Reinforcement Learning}
\newacronym{xrl}{XRL}{Explainable \glsentrylong{rl}}
\newacronym{ape}{APE}{Autonomous Policy Explanation}
\newacronym{llm}{LLM}{Large Language Model}
\newacronym{ret}{RET}{Remote Electrical Tilt}
\newacronym{rag}{RAG}{Retrieval-Augmented Generation}
\newacronym{sql}{SQL}{Structured Query Language}
\newacronym{dt}{DT}{Decision Tree}
\newacronym{kpi}{KPI}{Key Performance Indicator}
\newacronym{dqn}{DQN}{Deep Q-Network}
\newacronym{sinr}{SINR}{Signal-to-Interference-plus-Noise-Ratio}
\newacronym{rsrp}{RSRP}{Reference-Signal-Received-Power}
\newacronym{ue}{UEs}{User Equipments}
\newacronym{bs}{BS}{Base Station}
\newacronym{dnn}{DNN}{Deep Neural Network}
\newacronym{mp}{MP}{Multi-prompt Parsing}
\newacronym{ml}{ML}{Machine Learning}
\newacronym{dl}{DL}{Deep Learning}
\newacronym{xgboost}{XGBoost}{Extreme Gradient Boosting}
\newacronym{mdp}{MDP}{Markov Decision Process}
\newacronym{cbs}{CBS}{Clustering-Based Summarizer}
\glsdisablehyper

\usepackage{authblk}

\author[1,2]{Ahmad Terra}
\author[1,3]{Mohit Ahmed}
\author[1,2]{Rafia Inam}
\author[4]{Elena Fersman}
\author[2]{Martin Törngren}

\affil[1]{%
\centering
Ericsson Research, Ericsson AB, Stockholm, Sweden\par
(\texttt{ahmad.terra@ericsson.com}, \texttt{mohit.ahmed@ericsson.com}, \texttt{rafia.inam@ericsson.com})
}

\affil[2]{%
Industrial Engineering and Management, KTH Royal Institute of Technology, Stockholm, Sweden\\
(\texttt{terra@kth.se}, \texttt{raina@kth.se}, \texttt{martint@kth.se})
}

\affil[3]{%
Faculty of Science and Technology, Uppsala University, Uppsala, Sweden\\
(\texttt{mohit-uddin.ahmed.2400@student.uu.se})
}

\affil[4]{%
Global AI Accelerator, Ericsson Inc., Santa Clara, CA 95054, US\\
(\texttt{elena.fersman@ericsson.com})
}

\begin{document}
\title{Textual Explanations and Their Evaluations for Reinforcement Learning Policy}

\twocolumn[
\maketitle

\begin{abstract}
Understanding a \gls{rl} policy is crucial for ensuring that autonomous agents behave according to human expectations. This goal can be achieved using \gls{xrl} techniques. Although textual explanations are easily understood by humans, ensuring their correctness remains a challenge, and evaluations in state-of-the-art remain limited. 
We present a novel \gls{xrl} framework for generating textual explanations, converting them into a set of transparent rules, improving their quality, and evaluating them. Expert's knowledge can be incorporated into this framework, and an automatic predicate generator is also proposed to determine the semantic information of a state.
Textual explanations are generated using a \gls{llm} and a clustering technique to identify frequent conditions. 
These conditions are then converted into rules to evaluate their properties, fidelity, and performance in the deployed environment. 
Two refinement techniques are proposed to improve the quality of explanations and reduce conflicting information.
Experiments were conducted in three open-source environments to enable reproducibility, and in a telecom use case to evaluate the industrial applicability of the proposed \gls{xrl} framework. 
This framework addresses the limitations of an existing method, \glsentrylong{ape}, and the generated transparent rules can achieve satisfactory performance on certain tasks.
This framework also enables a systematic and quantitative evaluation of textual explanations, providing valuable insights for the \gls{xrl} field.

\end{abstract}

\begin{IEEEkeywords}
Explainable reinforcement learning, textual explanations, rule extraction,  explanation evaluation.
\end{IEEEkeywords}
\vspace{1em}
]

\section{Introduction}
\label{sec:introduction}

\gls{ai}, and especially \gls{dl} algorithms, are becoming more and more adopted in industry, increasing the need for transparency and explainability given the black-box nature of \gls{dl} algorithms. 
To address this need, \gls{xai} contributes by explaining how an \gls{ai} system arrives at a particular decision. 
As this is crucial for building user confidence and complying with \gls{ai} regulations, \gls{xai} adoption is also increasing in industrial applications~\cite{ahmed_2022_fromai2xai}.
A subfield of \gls{xai} is \gls{xrl}, which focuses mainly on understanding the \gls{rl} policy (a function that maps a state to an action). 
\gls{ape}~\cite{hayes_2017_improvingrobotcontroller} is an \gls{xrl} method that is suitable for non-\gls{ai} experts because the explanation is presented in a natural language format. 

Studies suggest that textual explanations are preferred over graphical ones regardless of expertise~\cite{malandri_2023_convxai}, but they can also cause over-reliance on \gls{ai}~\cite{he_2025_isconvxaiallyouneed}.
\gls{ape} requires fixed templates that lead to rigid interactions. On the other hand, \glspl{llm} have enabled humans to interact with machines flexibly without following a strict template. 
In \gls{xai}, \glspl{llm} have been used to compile attributive explanations through \gls{rag}~\cite{tekkesinoglu_2024_fromfi2nlpusingllm}. A recent study indicates that \glspl{llm} can improve \gls{xai} usability due to their narrative nature~\cite{zytek_2024_llms4xai}. 

Another key to enable human trust in \gls{ai} is the evaluation of explanations.
In textual explanations, especially when \glspl{llm} are employed, some evaluations focus on how they are presented \cite{zytek_2024_llms4xai}.
In a more traditional \gls{xai} setting, ground-truth explanations are rarely available. 
Alternatively, explanations can also be evaluated by involving humans, although this remains challenging as it is prone to bias \cite{kadir_evaluation_2023}.

In this article, we propose a novel framework to explain the behavior of an \gls{rl} agent by generating and evaluating its textual explanations. This framework focuses on producing a list of state conditions relevant to actions with consistent categorical values. 
In this framework, a user can query the explanations and receive them in a natural-language format.
The explanations are evaluated on the basis of how representative they are relative to the obtained data and the given task. 
The main contributions are as follows:
\begin{itemize}
\item An implementation of an \gls{llm} to interpret queries about an \gls{rl} policy.
\item A method to summarize an \gls{rl} policy for textual explanations. This includes techniques to define, extend, and refine predicates in the discretization of data.
\item A method for extracting rules from textual explanations and evaluating them to assess their correctness.
\item Two techniques to improve explanation quality.
\item A framework that integrates all the above points in three different open-source environments and an industrial-telecom application. 
\end{itemize}

\section{Related Work}
\label{sec:related_works}

\gls{ape} \cite{hayes_2017_improvingrobotcontroller} analyzes historical data to generate explanations of conversation-like interactions between humans and an \gls{rl} policy. This method can be used to generate meaningful counterfactuals and filter attributive explanations \cite{hefny_2025_comprehensiveexplanations}.
In operation, predicate functions are used to convert numerical values into textual information.
It has a limitation in scaling up to large predicate sets due to the employed Quine-McCluskey algorithm \cite{mccluskey_1956_minimization}. 
The computational complexity of the algorithm increases exponentially with the predicate size. In addition, when all possible states are present in an action, there is no indication of the condition under which the \gls{rl} agent would take that action. 
Furthermore, \gls{ape} can generate duplicate explanations when similar conditions appear for different actions. 
Our proposed method addresses these limitations by presenting frequent conditions and suppressing duplicates with linear growth in computation relative to the predicate size.

Generating textual explanations can also be done by adding explainer modules to the \gls{dnn} model, as done in InterpNET\cite{barratt_2017_interpnet} and EPReLU-CSGNN-MALSTCAM~\cite{sheela_2025_efficient}. In these works, the classifier module was initially trained, followed by training the \gls{dnn} explainer module. Although a higher accuracy was achieved, it contradicted the spirit of explainability in minimizing the black-box properties of the model. 
A more direct method was presented in \cite{garciamagarino_2019_hcaitwai4iot}, where the most relevant nodes were identified and presented in a textual explanation using a predefined template. 
Alternatively, Poli et al. proposed a method to generate a textual explanation for semantic image annotation \cite{poli_2021_textual_semantic_annotation}, also using a predefined template. %human evaluation is involved
Similar to these methods, we use a predefined template, but instead of generating a local explanation as used in the aforementioned methods, our method combines conditions covering the global or cohort scope.

\glspl{llm} can process and generate textual information in a flexible manner with a less rigid format. An implementation of an \gls{llm} to explain an \gls{ai} model was presented in \cite{tekkesinoglu_2024_fromfi2nlpusingllm}. In that work, \gls{rag} was used to collect relevant information about the data instances being explained, including their attribution values. More thorough frameworks have been proposed in 
TalkToModel~\cite{slack_explaining_2023} and CoXQL~\cite{wang_2024_coxql}, where multiple \gls{xai} techniques are integrated to handle different types of explanations. They focused on the correctness of the generated parsing command for collecting data from the database and converted the results into text using an \gls{llm}. 
Our proposed method uses an \gls{llm} as a question interpreter, and our focus is on generating textual explanations from the parsed data. 

In the \gls{xrl} field, NavChat~\cite{trigg_2024_nlp4navigation} used an \gls{llm} to justify and explain the decisions made by the \gls{rl} agent with DeepSHAP~\cite{lundberg_2017_shap} to identify impactful features. In addition, HighwayLLM~\cite{yildirim_2025_highwayllm} used an \gls{llm} to explain the \gls{rl} policy and ensure safety. 
In the telecom field, Ameur et al.~\cite{ameur_2024_llmxrl6g} used an \gls{llm} to explain the \gls{rl} agent controlling a network slicing problem. 
Although most previous methods are applied to specific applications, ours is application-agnostic, as exemplified in several environments, including an industrial application.

In the \gls{xai}-rule-extraction field, a direct extraction of a \gls{dnn} model is carried out through a \gls{dt} \cite{mahbooba_explainable_2021}. 
The FRBES~\cite{aghaeipoor_2023_frbes} extracts fuzzy rules from a \gls{dnn}, which employs an attributive explainer to prioritize important features for constructing the rules. 
CGX~\cite{hemker_2023_cgxplain} uses a decomposition method to extract rules from the hidden representation of a \gls{dnn} model, where multiple objectives can be defined to fit the explainability needs. 
In \gls{rl}, Engelhardt et al. extracted rules from trained agents using a \gls{dt} \cite{engelhardt_sample_based_2023}. 
Although they used rules as explanations, ours is the opposite, i.e., the explanation is converted into rules with consistent threshold values. 
This technique bridges textual explanations and their evaluation.

Most evaluation metrics in \gls{xai} are proposed for attributive explanations. 
Researchers have proposed many metrics with similar terms, but they sometimes have significantly different formulations \cite{subhash2022what}. Challenges such as inconsistent findings or oversimplification were also reported in \cite{sokol_2024_xaieval}.
In \gls{xai} with rule extraction, information power is proposed to evaluate the explanation involving non-expert users during the evaluation~\cite{matarese_how_2025}. %also needs deterministic system 
In textual explanations with \glspl{llm}, evaluations often focus on the correctness of parsing the required information \cite{slack_explaining_2023,wang_2024_coxql}.  Our experiment and evaluation follow the suggestions of \cite{sokol_2024_xaieval}, which addresses real-world telecom use cases while being evaluated in a rather simplified scenario. Instead of evaluating the explanations in their local scope, we analyzed the coherence of the explanations to ensure consistency. 

\section{Proposed Framework}
\label{sec:proposed_framework}

In \gls{rl}, an agent learns through interactions with an environment — an external system that provides a state that represents its condition ($s$), receives an action that influences it ($a$), and returns a reward as feedback ($r$). 
The states are typically represented by the input features with continuous values, and the predicate functions ($P$) can be used to discretize and represent their semantic information. 
In this work, it is assumed that the \gls{rl} agent has learned its policy ($\pi$) and stored the replay data consisting of a set of states (S), actions (A), and rewards (R) in an SQLite database. 

\begin{figure}[h]
    \centerline{\includegraphics[width=0.97\columnwidth]{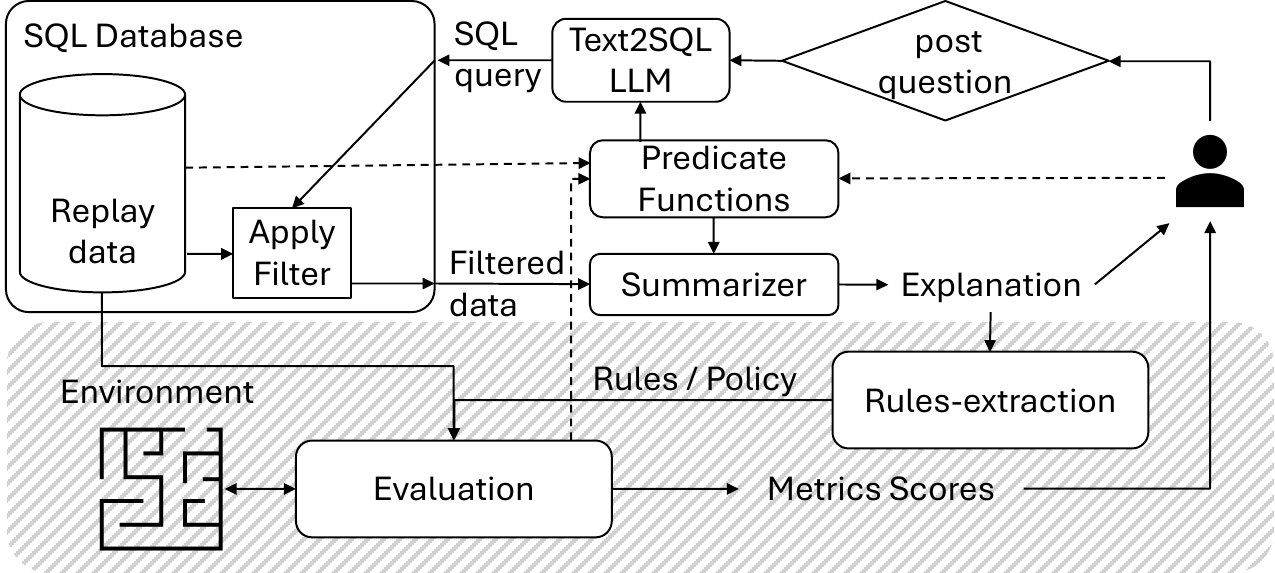}}
    \caption{Overall framework of the proposed method, where the top half illustrates the explanation generation process, while the shaded area represents evaluation using rule extraction.}
    \label{fig:overall_framework}
\end{figure}

We propose a framework to explain an \gls{rl} policy by summarizing the relationships between states and actions in a textual format. It can interactively answer questions such as "\textit{when will you do \textless action\textgreater?}" or "\textit{what will you do on \textless condition\textgreater?}". A condition ($c$) is a set of partial or complete logical statements that construct a discrete state. 
The overall framework is illustrated in \cref{fig:overall_framework}, where it begins with a user asking a question, which is then converted into a \gls{sql} query via a text-to-SQL component to filter the data. The filtered data are then summarized using our proposed method to generate textual explanations. These explanations are then converted into rules for evaluation using replay data and the environment. The details of each stage are presented in the following subsections.

\subsection{LLM as Explainer Interface}
The \gls{llm} is used to interpret questions about the \gls{rl} policy to convert them into an \gls{sql} query \cite{shi_survey_2025} to retrieve the relevant data. The user’s question is accompanied by a few-shot learning to prompt the \gls{llm} to ensure the correct \gls{sql} query format. Once generated, the query is then executed on the database, and the filtered data is passed to the summarizer. 
The \gls{llm} serves solely as the interface for query interpretation, while the summarizer performs data aggregation; the two are decoupled to maintain transparency and control.

\subsection{Clustering-Based Summarizer (CBS)}
The result of the above step is expected to be a long list of data instances relevant to the query. Since these are numerical, we propose the \gls{cbs} to summarize and present them in a textual format.

\subsubsection{Predicate Function}
\label{subsec:predicate_function}
Let us define $p$ as a predicate function that takes a numerical state ($s \in S$) and returns a set of logical values or levels indicating how the state complies with $p$. In \gls{ape}, $P$ is a set of Boolean predicate functions that discretize the state. 
Let $f$ be the feature name of a complete feature set $F$ ($f \in F$), thus $p^f(s)$ is the predicate function for the feature $f$. With binary functions, the predicate can be formalized as $p^f(s) = \mathbb{I}(s^f > l^f)$ where $l$ is a limit/threshold and $\mathbb{I}$ is an indicator function that returns a Boolean value. Thus, each input feature is converted into a true/false or high/low value. 

In this framework, an expert can manually define the predicate functions. 
Alternatively, statistical methods can be used, such as choosing the median or a certain quantile value of the data as a threshold. 
However, this can be problematic if the data distribution has high density around it because many similar instances would be discretized to different values.
Therefore, we propose a technique to determine the predicate limits, as shown in \cref{alg:gini_predicate}. Here, we slice the data by feature names ($S^f$) to train a \gls{dt} with a maximum of $n_{cat}$ leaf nodes. 
$L$ is the set of limits obtained from the splitting values in the \gls{dt}.

\begin{algorithm}[h]
    \caption{Gini-based Predicate Limit Generation}
    \label{alg:gini_predicate}
    \begin{algorithmic}[1]
        \REQUIRE States: $S$; Actions $A$; Feature names $F$; Number of categories $n_{cat}$;
        \STATE $L \gets \emptyset$

        \FOR{$f \in F$}
            \STATE $X \gets S^f; Y \gets A$ %L^f \gets \emptyset; 
            \STATE $DT^f \gets \mathrm{train\_DT(X, Y, n_{cat})}$ 
            \COMMENT{Train a decision tree with maximum of $n_{cat}$ leaf nodes}
            \STATE $\Theta \gets \mathrm{get\_splitting\_values(DT^f)}$
            \STATE $L^f \gets L^f \cup \Theta$ \COMMENT{append threshold(s) to $L^f$}

        \ENDFOR
        \RETURN $L$
    \end{algorithmic}
\end{algorithm}

\subsubsection{Predicate Levels}
When a binary function is used, each predicate function outputs two levels of information. In a binary system, categorizing a feature into "low", "medium", and "high" requires three different functions and generates three digits/elements to represent them. Thus, we propose to use a categorical predicate formulation in \cref{eq:categorical_predicate} to simplify this process. In this way, different levels of the same feature can be represented using a single digit/value. 
\begin{equation}
	P_f(s) =
	\begin{cases}
		-1.0, & s^f \le L^f_{0} \\ 
		\frac{2}{n_{cat} -1 } -1 , & L^f_{0} < s^f \le L^f_{1} \\
        \dots, & \vdots \\
		\frac{2 \times (n_{cat} - 2)}{n_{cat} -1 } -1 , & L^f_{n_{cat}-2} < s^f \le L^f_{n_{cat}-1} \\
        1.0, & L^f_{n_{cat}-1} < s^f
	\end{cases}
	\label{eq:categorical_predicate}
\end{equation}

\subsubsection{Clustering Summarization}
We cluster the discretized data using the K-means algorithm to collect similar instances into one group and use the elbow method to determine the number of clusters. The K-means algorithm is used because the categorical values contain magnitude information for each relevant predicate. 
In each cluster, unique instances are computed and sorted in descending order. As a cluster may have outliers or infrequent instances, we set an inclusion threshold $\theta \in (0, 1]$ to filter these out.
We collect the frequent discrete states in $Q$ and iterate over all predicates $P$. Let $Q^p$ be the set of discrete values of $Q$ for the predicate $p$, $\psi^p_v$ be the tuple of the predicate function $p$ and the value assigned to it ($\psi^p_v = \langle p, p(s) \rangle$). When there is one unique discrete predicate value in the cluster ($\bigl|\{Q^p\}\bigr| = 1$), $\psi^p_v$ is added to the condition $c_k$. 
After collecting the conditions from each cluster, the summary is returned as a list of conditions $\{c_0, c_1,...,c_k\} \in C$ as presented in \cref{alg:clustering_summarization}. 
\begin{algorithm}[h]
    \caption{Clustering-based Summarization (CBS)}
    \label{alg:clustering_summarization}
    \begin{algorithmic}[1]
        \REQUIRE States $S$; Predicate funcs $P$; Inclusion thresh $\theta$;
        \STATE $C \gets \emptyset$

        \STATE $D \gets P(S)$ \COMMENT{Discretize the filtered states}

        \STATE $K \gets \mathrm{KMeans\_clustering(D)}$

        \FOR{$k \in K$} 
            \STATE $n \gets |D_k|$ \COMMENT{Count the number of $D_k$}

            \STATE $U_k \gets \mathrm{count\_unique(D_k)}$ \COMMENT{Collect frequency of $D_k$}
            \STATE $U^{\downarrow}_k \gets sort(U_k)$ \COMMENT{Sort in descending order}
            
            \FOR{$i = 1,\dots,|U_k^{\downarrow}|$}
                \STATE $M[i] \gets \sum_{j=1}^{i} U_k^{\downarrow}[j]$  \COMMENT{cumulative sum}
                \IF{$M[i] \geq \mathrm{int(\theta \times n)}$}
                \STATE $m \gets i$ \COMMENT{last index within threshold}
                \STATE \textbf{break}
                \ENDIF
            \ENDFOR
            \STATE $Q \gets \{\,U_k^{\downarrow}[1],\dots,U_k^{\downarrow}[m]\,\}$  
                \COMMENT{keep top-n $U_k^{\downarrow}$}
            \STATE $c_k \gets \emptyset$
            \FOR{$p \in P$}
                \IF{$\bigl|\{Q^p\}\bigr| = 1$} %\COMMENT{if only one unique value}
                    \STATE $c_k \gets c_k \cup \langle p, Q^p_0 \rangle $ \COMMENT{append $\psi_p^Q$ to $c_k$}
                \ENDIF
                
            \ENDFOR
            \STATE $C \gets C \cup c_k$ \COMMENT{append $c_k$ to $C$}

        \ENDFOR
        \RETURN $C$
    \end{algorithmic}
\end{algorithm}

\subsubsection{Textual Explanation}
Converting a condition into a textual explanation follows the defined predicates. For example, $p_1$, $p_2$ and $p_3$ are binary predicate functions that discretize features $f_1$, $f_2$, and $f_3$, respectively. 
If the resulting vector is $c=(-1,1,-1)$, then the textual information is "$f_1$ \textit{is low AND} $f_2$ \textit{is high AND} $f_3$ \textit{is low}"\label{example:textual_explanation}. When multiple conditions are identified, they are concatenated with the "OR" operator. 
In categorical discretization, textual levels can be extended to accommodate more values, for example, when $n_{cat}=5$, textual levels can be defined as "Very Low", "Low", "Medium", "High", and "Very High". When multiple levels are identified for the same predicate, we compress them by combining them into a single criterion, such as "Above Low" or "Below Medium".
We use a template rather than an \gls{llm} to convert a condition into a textual explanation to save computational resources, as \glspl{llm} are not designed to process numerical information. 
Additionally, we experimented with \verb|DeepSeek|-\verb|14B|, \verb|DeepSeek|-\verb|32B|, and \verb|Mixtral|-\verb|8x7B|-\verb|Instruct| to convert the filtered data into text, but they discarded a fraction of the information and had flaws in presenting the observations. 

\subsection{Rules Extraction}
\label{subsec:rules_extraction}

As textual explanations imply a correlation between conditions and actions, humans may assume that when a condition is met, the same action will follow.
We propose a method for converting textual explanations into rules, presented in \cref{alg:rules_extraction}. The first step in this process is to collect the conditions for each action. However, the same condition may lead to different actions due to the discretization process. Thus, further information is needed to convert textual explanations into rules in which the occurrence ($o$) of a condition can be used to determine an action. For example, if action $a$ is taken more frequently than the others under the same conditions, then the generated rules will select this action. 

\begin{algorithm}[h]
    \caption{Rules Extraction from Explanations}
    \label{alg:rules_extraction}
    \begin{algorithmic}[1]
        \REQUIRE States: $S$; Actions: $A$; Predicate functions $P$; Weight function $w$; Inclusion threshold $\theta$; 

        \STATE $\mathcal{R} \gets \emptyset$

        \FOR{$a \in \mathrm{unique(A)}$}
        \STATE $S_a \gets \{\, s \in S \mid \pi(s) = a \,\}$
            \STATE $C_a \gets \mathrm{Summarize(S_a, P, \theta)}$ \COMMENT{\cref{alg:clustering_summarization} or \gls{ape}}
            \STATE $\mathcal{R}_a \gets \emptyset$
            \FOR{$c_a \in C_a$} 
                \STATE $o_{c,a} \gets \mathrm{w(S, A, c_a, a)}$
                \STATE $\mathcal{R}_a \gets \mathcal{R}_a \cup \langle c_a, o_{c,a} \rangle$
            \ENDFOR
            \STATE $\mathcal{R} \gets \mathcal{R} \cup \mathcal{R}_a$
        \ENDFOR
        
        \RETURN $\mathcal{R}$
    \end{algorithmic}
\end{algorithm}

The second step is to assign a weight that indicates the occurrence of every condition in all actions. This is achieved by computing the state occurrence from the replay data. Let us define the total number of instances as $N_{total}$, the number of instances in a condition as $N_c$, the number of instances in an action as $N_a$, and the number of instances in a condition with the same action as $N_{ca}$. We then examine the following formulas to determine the best weighting function ($w$) to select the action: $w_1 = \frac{N_{ca}}{N_a}$, $w_2 = \frac{N_{ca}}{N_c}$, $w_3 = \frac{N_{ca}}{N_{total}}$, and $w_4 = \frac{N_{ca}}{N_a}\times \frac{N_{ca}}{N_{total}}$.
The percentage of conditions that appear over all data with the same action is indicated by $w_1$, while $w_2$ is the percentage of an action taken under the same condition. $w_3$ computes the percentage of the same condition-action over all data and $w_4$ is the combination of $w_1$ and $w_3$.

Finally, each condition ($c_a$) is coupled with its occurrence ($o_{c,a}$) to construct a rule for action $a$ ($\mathcal{R}_{c,a}(s) = \mathbb{I}(P(s) = c_a).o_{c,a}$). In other words, this can be interpreted as "\textit{when condition} $c_a$ \textit{is met, action $a$ is taken with} $o_{c,a}$ \textit{occurrences}". To utilize the generated rules $\mathcal{R}$, we use the following formula in \cref{eq:action_selection} to select the best action ($a^*$), where $C_a$ is the set of conditions for an action:  
\begin{equation}
	\label{eq:action_selection}
	a^* \;=\;\operatorname*{arg\,max}_{a\in\mathcal A}
	\sum_{c\in\mathcal C_a} 
	\mathbf{\mathcal{R}_{c,a}(s)}%\{\,\text{condition }c\text{ is satisfied by }a\}\,,
\end{equation}

In the generated rules, a state may not be covered in any action due to the removal of infrequent occurrences or new conditions that have not been observed in the replay data. To solve this, we propose an approximation function (\cref{alg:state_condition_approximation}) to cover a state that is not present in any condition/rule and determine its action. 
Euclidean distance is used as the data are in a normalized range.
Using this approach, similar textual explanations, including \gls{ape} explanations, can be converted into rules. If the evaluation results are satisfactory, the black-box agent can be replaced by these transparent rules.
\begin{algorithm}[h]
    \caption{State-Condition Approximation}
    \label{alg:state_condition_approximation}
    \begin{algorithmic}[1]
        \REQUIRE state $s$; Conditions $C$; Predicates $P$; Features $F$
        \STATE $d_{min} \gets \infty $
        \FOR{$c \in C$}
            \STATE $m_{c} \gets \mathrm{median(c, P)}$ \COMMENT{bin centers of $c$ given $P$}
            \STATE $d \gets \sqrt{\sum_{f \in F}(m^{f} - s^{f})^2}$ \COMMENT{Euclidean distance}
            \IF {$d < d_{min}$}
                \STATE $d_{min}, c_{min} \gets d, c$
            \ENDIF
        \ENDFOR
        \RETURN $c_{min}$
    \end{algorithmic}
\end{algorithm}

\subsection{Explanation Evaluation}
Textual explanations contain information that is easily understood by humans.
The length of an explanation plays an important role in understanding it. Formally, the length of the explanation is defined by the number of conditions generated ($E_{\text{len}} = |C|$). Although longer explanations may exhaust one's mental capacity, they capture more detailed information than the shorter ones. 

As mentioned in \cref{subsec:rules_extraction}, the same condition may appear in multiple actions. As this may lead to ambiguity, we count the number of duplicated conditions as part of our evaluation ($E_{\text{duplicate}} = \left| \left\{ C_i \;\middle|\; \exists\, a_1 \neq a_2,\; C_i \in C^{a_1} \wedge C_i \in C^{a_2} \right\} \right|$). These two metrics are categorized as explanation properties, as they directly assess the explanations without comparing them with the underlying replay data used to generate them.

Furthermore, we use the replay data for evaluation, where we count the states that cannot be mapped to any condition on any action, $\bigl( E_{\mathrm{approx}} = \frac{1}{|S|} \sum_{s \in S} 1\bigl(\forall a,\; P(s) \notin C^a\bigr)\bigr)$. This metric can be considered a coverage metric that shows how many states are not covered by the generated explanations/rules.

The states in the replay data are then used to infer the actions of the proposed rule extraction. 
In this case, the actions in the replay data can be considered as the ground truth of the policy being explained. Consequently, we can determine the fidelity of the explanations by comparing them using supervised learning metrics such as accuracy ($E_{\text{acc}}$), recall ($E_{\text{rec}}$), and F1 ($E_{\text{F1}}$).

In \gls{rl}, the reward is an indicator of how an agent performs in the environment. Since the aim of this work is to explain the \gls{rl} agent's behavior, we can also evaluate the performance of the explanation by converting it using the proposed rule extraction method. This is done by deploying the rules in the same environment in which the \gls{rl} agent was previously trained. We collect three metrics in doing so, which are: cumulative reward in an episode ($E_{\text{CR}}=\sum_{t=0}^{T} r_t$), total time steps in an episode ($E_{\text{TS}} = T$), and average reward per time step ($E_{\text{AR}}=\frac{E_{\text{CR}}}{E_{\text{TS}}}$), where $r$ is the reward and $T$ is the total time steps. Note that the reward is a quantitative measure that can be replaced by any other numerical indicator, such as cost savings, \gls{kpi}, energy consumption/savings, etc. Thus, the proposed evaluation method can be used to indicate the cost/compromise if one chooses to use the extracted transparent agent. This performance evaluation enables better trade-off analysis between black-box and transparent agents.

\subsection{Predicate Refinement}
\label{subsec:predicate_refinement}
In \cref{subsec:predicate_function}, we presented a method for defining the predicate thresholds based on the Gini score of the filtered data. It initializes the predicates based only on the data, neither considering the generated explanations nor the rules. In this subsection, we introduce methods for refining the predicate definitions by considering the generated rules. We propose a technique to reduce $E_{\text{duplicate}}$ with coherent and less conflicting information among actions. Alternatively, we also propose a technique to improve the fidelity (F1) of the explanations. 

\begin{algorithm}[h]
    \caption{Minimize Duplicated Conditions Among Actions}
    \label{alg:minimize_duplicates}
    \begin{algorithmic}[1]
        \REQUIRE States: $S$; Actions $A$; F1 score $f1$; Number of categories $n_{cat}$; Duplicated Conditions $C$; Predicate Limits $L$; Max Recommendations $m$; Iteration Budget $b$;
        \STATE $L_{update} \gets L; L_{init} \gets L; i \gets 0$
        \STATE $L_{c} \gets \emptyset$ \COMMENT{collection of proposed limits}
        \STATE $D \gets \emptyset$ \COMMENT{tracking count of duplicate}

        \WHILE{$i < b$}
        \STATE $F \gets \emptyset$ \COMMENT{tracking F1 score}
        \FOR{$c \in C$}
            \STATE $S_c, A_c \gets \mathrm{filter(S, A, c)}$ \COMMENT{filter data that satisfy $c$}
            \STATE $DT_f \gets \mathrm{train\_DT(S_c, A_c, n_{cat})}$ 
            \COMMENT{Train a decision tree with maximum of $n_{cat}$ leaf nodes}
            \STATE $\Theta \gets \mathrm{get\_thresholds(DT_f)}$
            \STATE $\Theta^\uparrow  \gets \mathrm{sort(\Theta)}$ \COMMENT{sort from root to the farthest leaf}

            \FOR{$\theta \in \Theta^\uparrow$}
                \STATE $j \gets \mathrm{argmin(abs(L^f_{update} - \theta^f))}$ \COMMENT{nearest limit}
                \STATE $L^{f,j}_{update} \gets \theta^f$
                \STATE $d, f1 \gets \mathrm{P.apply(L_{update})}$ \COMMENT{get duplicate \& f1}
                \STATE $D \gets D \cup d; L_{c} \gets L_{c} \cup L_{update}; F \gets F \cup f1$ %\COMMENT{}
                \STATE $L_{update} \gets L_{init}$
            \ENDFOR
            \IF{$\mathrm{max(F)} > f1$}
                \STATE $L_{init} \gets L_{\mathrm{max(F)}}; f1 \gets \mathrm{max(F)}$
            \ENDIF

        \ENDFOR
        \STATE $i \gets i + 1$
        \ENDWHILE
        \STATE $k \gets \mathrm{argmin(D)}$
        \RETURN $L_{candidate, k}$
    \end{algorithmic}
\end{algorithm}

The technique for minimizing duplicate conditions ($E_{\text{duplicate}}$) is presented in \cref{alg:minimize_duplicates}. It iterates over the duplicated conditions (line 7) and then builds a \gls{dt} model from the filtered data to obtain new threshold values ($\Theta$). 
Each value is then evaluated using the duplicates and F1 score. If the proposed change increases the F1 score, then the updated threshold is used in the next iteration. The updated thresholds are collected during these iterations, and the one with the lowest duplicates is returned.

\begin{algorithm}[h]%[htbp]
    \caption{Maximize F1 Score}
    \label{alg:maximize_f1}
    \begin{algorithmic}[1]
        \REQUIRE States: $S$; Feature Names $F$; Predicate Limits $L$; Adjustment rate $\alpha$; Iteration Budget $b$;

        \STATE $f1 \gets \mathrm{P.apply(L)}$
        \STATE $f1_{1} \gets f1; f1_{2} \gets f1$

        \FOR{$f \in F$}
            \FOR{$i \in L^f$}
                \STATE $X \gets S^f$ \COMMENT{collect feature f values}
                \IF{$\mathrm{len(L^f) > 1}$} %\COMMENT{multiple levels of predicate}
                    %\STATE $X \gets L^{f}_{i-1} \leq S^f < L^{f}_{i+1}$
                    \STATE $X \gets \{ s_f \in S_f | L^f_{i-1} \leq s_f < L^f_{i+1} \} $
                \ENDIF
                \STATE $\theta \gets L^{f}_{i}; t \gets False; j \gets 0$ 
                \WHILE{$(\not t) $AND$ (j < b)$}
                    
                    \STATE $k \gets 1$
                    \WHILE{$(f1_1 == f1) $AND$ (k < b) $}
                        \STATE $\theta_1 \gets \theta - k \times \alpha \times (\theta - \mathrm{min(X)})$
                        \STATE $f1_1 \gets \mathrm{P.apply(\theta_1)}; k \gets k + 1$ 
                    \ENDWHILE

                    \STATE $k \gets 1$
                    \WHILE{$(f1_2 == f1) $AND$ (k < b) $}
                        \STATE $\theta_2 \gets \theta + k \times \alpha \times (\mathrm{max(X)} - \theta)$
                        \STATE $f1_2 \gets \mathrm{P.apply(\theta_2)}; k \gets k + 1$ 
                    \ENDWHILE

                    \STATE $m \gets \mathrm{argmax(\{f1, f1_1, f1_2\})}$
                    \IF{$m == 0$}
                        \STATE $t \gets True$
                    \ENDIF
                    \STATE $L^{f,i} \gets \theta_m; f1 \gets f1_m$
                    \STATE $j \gets j + 1$
                \ENDWHILE
            \ENDFOR
        \ENDFOR
        
        \RETURN $L$
    \end{algorithmic}
\end{algorithm}

The technique for maximizing the F1 score is proposed in \cref{alg:maximize_f1}, in which the predicate thresholds are adjusted around their original values. It starts by computing the F1 score of the current setup (L) and then iterates over all features F and each threshold within it ($L^f$). If the predicate function is nonbinary (level $\ge 2$), it collects the data that will be affected by the threshold ($L^{f,i}$; line 7). For these selected data, we then try to decrease (line 13) or increase (line 18) the threshold and collect the F1 scores. The threshold with the highest F1 value is then used in the next iteration (lines 21-25).

\section{Experiments and Evaluations}
\label{sec:experiments}

In this section, we first present the experimental setup. 
We then present the results of the comparison of \gls{ape} and \gls{cbs}, followed by experiments with varying parameters in \gls{cbs}. 

\subsection{Experiment Setup}
In this work, we used three open-source environments from the Gymnasium\footnote{\href{https://gymnasium.farama.org}{https://gymnasium.farama.org}} library: LunarLander-v2, MountainCar-v0, and CartPole-v1. We used pre-trained \gls{dqn} agents\footnote{\href{https://github.com/DLR-RM/rl-trained-agents/}{https://github.com/DLR-RM/rl-trained-agents/}} for these environments to ensure the optimal performance of the agent and enable reproducibility. %stable_baselines3 import DQN

\begin{figure}[h]
    \centerline{\includegraphics[width=0.6\columnwidth]{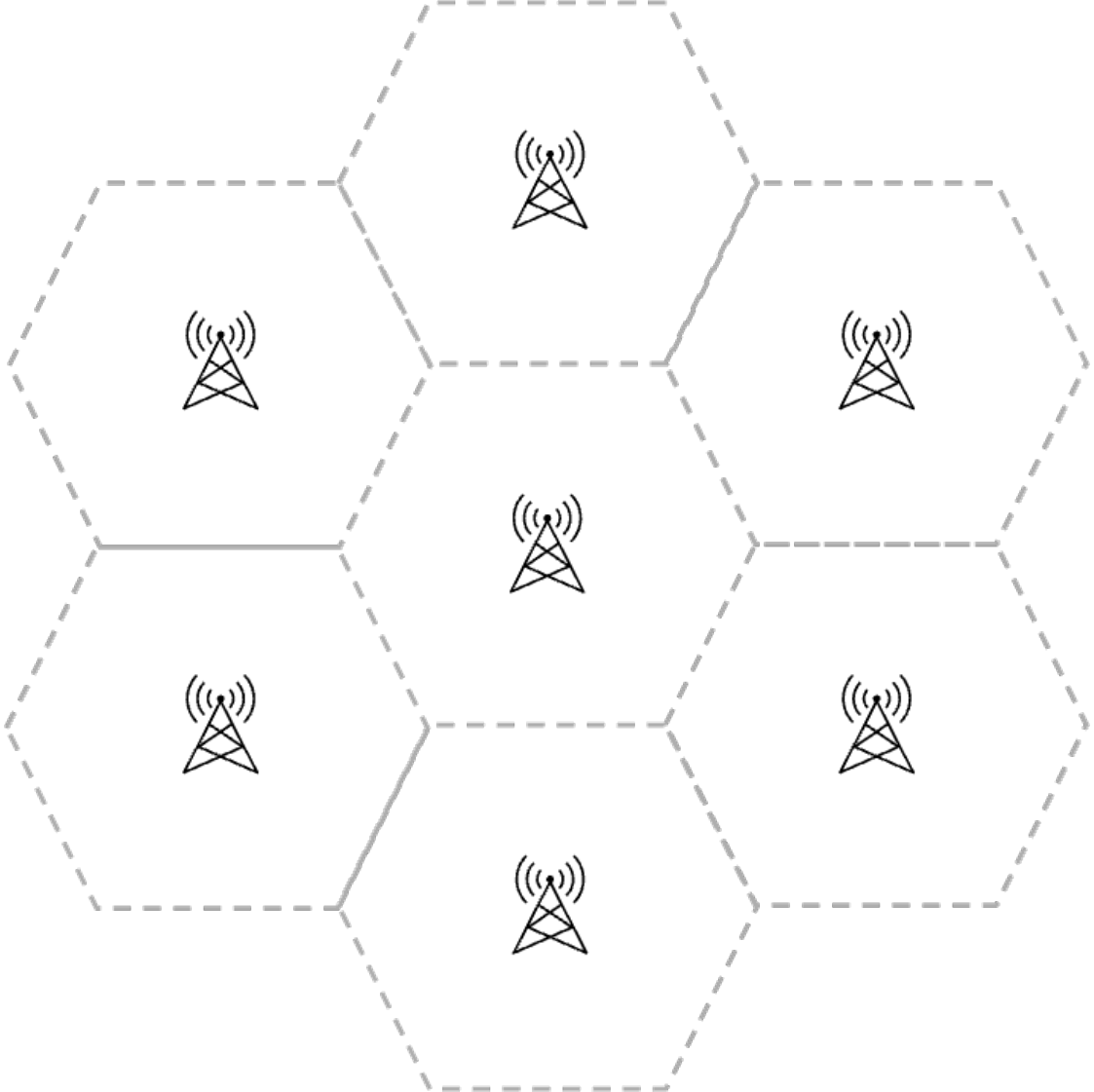}}
    \caption{Layout of base stations used in \gls{ret} use case.}
    \label{fig:ret}
\end{figure}

Additionally, we also experimented with an industrial use case of a proprietary \gls{ret} simulated environment, where the goal is to optimize its \gls{kpi}s by controlling the tilt of the base-station antennas. 10,000 \gls{ue} are uniformly distributed across the environment, and seven \glspl{bs} are configured in a hexagonal shape, as shown in \cref{fig:ret}. The state comprises \gls{ue} interference, \gls{sinr}, \gls{rsrp}, throughput, and antenna tilt angle. The actions are discrete: $1^\circ$ up-tilt, no tilt, and $1^\circ$ down-tilt.
The reward is formulated as $R = 0.6 \times \mathbb{I}(\text{RSRP} > -110 \text{dBm}) + 0.4 \times \mathbb{I}(\text{\gls{sinr}} > 13 \text{dB})$. The first part of the reward indicates the coverage of the signal, whereas the latter indicates the quality. Each antenna is controlled by an independent \gls{rl} agent that is trained using a \gls{dqn} with reward decomposition (DQNRD)~\cite{juozapaitis_2019_xrlrd} sharing the same policy. 

For the explainer interface, the \verb|DeepSeek|-\verb|R1|-\verb|Distill|-\verb|Llama|-\verb|70B| model \cite{deepseekai2025deepseekr1incentivizingreasoningcapability} with a temperature of $0.5$ is used.
The replay data was collected by running the trained agents in their environments until they reached 10,000 time steps and completed the final episodes. It starts with a seed equal to 0 and increases by 1 in every episode. This data is then used to generate explanations, extract rules, and evaluate them. For performance evaluation, we ran ten episodes for every set of experiments with the same seed numbers (0 to 9). 
This seed number was kept identical for all parameter configurations to allow for a valid comparison. 
The performance of \gls{ret} is only available in \cref{subsec:ret_results} because it took significant resources to run a single simulation.%

\subsection{Sample of the Explanations}
\begin{figure}[h]
    \centerline{\includegraphics[width=0.9\columnwidth]{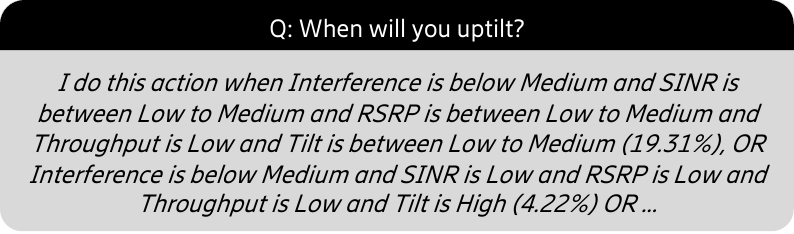}}
    \caption{Partially generated textual explanations.}
    \label{fig:explanations_samples}
\end{figure}

The proposed framework was built to summarize the observed data and present them in a textual format. A sample of the generated explanations in this experiment is shown in \cref{fig:explanations_samples}, in which we show only a partial explanation of the \gls{ret} use case to save space. Other types of questions, such as "\textit{What will you do when RSRP is low?}" can also be asked to generate explanations in a similar format. 

\subsection{APE vs \gls{cbs}}
In this subsection, we compare the evaluation of \gls{ape} with our proposed method, \gls{cbs}. We varied the weighting functions ($w$) and the predicate definitions for both methods. The inclusion thresholds were varied only for \gls{cbs} because \gls{ape} always includes the entire data ($\theta = 1.0$). The maximum number of clusters (k) in \gls{cbs} was set to 40. In this experiment, we used only two levels of predicates (Low and High) due to \gls{ape}'s limitation on binary predicates. The predicate functions for both methods were kept the same. 
The mean values of the metrics for each setup in this comparison are shown in \cref{tab:ape_vs_ccs}. 
Additionally, the values of reward-related scores ($E_{CR}$ and $E_{TS}$) are normalized by the highest value in each environment ($\mathrm{Norm(x) = x / max(abs(X_{env})) \times 100 \%}$) to fit the table width and ensure a fair averaging.

\begin{table}[h]
    \caption{Evaluation of APE (A) and \gls{cbs} (C) in different environments.} 
    \label{tab:ape_vs_ccs}
    \centering
    
\resizebox{\columnwidth}{!}{
\begin{tabular}{|c|r|r|r|r|r|r|r|r|}
\hline
 Env & \multicolumn{2}{|c|}{CartPole} & \multicolumn{2}{|c|}{LunarLander} & \multicolumn{2}{|c|}{MountainCar} & \multicolumn{2}{|c|}{RET} \\
 \hline
% Method & APE & CCS & APE & CCS & APE & CCS & APE & CCS \\
 Method & A & C & A & C & A & C & A & C \\
\hline
$E_{app}$  & - & 1.7 & 0.0 & 17.2 & - & 1.5 & 0.0 & 3.1 \\
$E_{len}$  & - & 13.6 & 56.9 & 24.4 & - & 7.6 & 16.8 & 17.9 \\
$E_{dup}$  & - & 3.8 & 1.9 & 2.3 & - & 2.6 & 0.0 & 3.7 \\
\hline
$E_{acc}$  & - & 73.2 & 42.8 & 41.4 & - & 75.1 & 72.0 & 72.6 \\
$E_{rec}$  & - & 73.2 & 45.5 & 36.4 & - & 59.6 & 62.4 & 57.3 \\
$E_{F1}$  & - & 72.3 & 35.1 & 28.9 & - & 54.0 & 60.2 & 52.4 \\
\hline
$E_{CR}$  & - & 17.4 & -69.2 & -35.2 & - & -83.3 & - & - \\
$E_{TS}$  & - & 17.4 & 38.5 & 28.3 & - & 83.3 & - & - \\
$E_{AR}$  & - & 1.0 & -1.8 & -1.2 & - & -1.0 & - & - \\
\hline
\end{tabular}
}
\end{table}

In the LunarLander and RET environments, there is no approximated state to decide the action ($E_{app}=0.0$) on \gls{ape} explanations, i.e., the generated rules cover any possible state. This is because the Quine-McCluskey algorithm is used to minimize the binary representation in a complete manner without discarding any binary state. However, \gls{ape} did not generate any explanation for the CartPole and MountainCar environments ($'-'$). All possible combinations of discrete states appeared in one or more action(s) in the replay data. Therefore, no rule or evaluation metric could be generated. 

The explanation length of \gls{ape} is significantly greater than \gls{cbs} for LunarLander, whereas it is slightly smaller for \gls{ret}. The duplicated conditions are also higher in the \gls{cbs} than in the \gls{ape} for both environments. Note that in LunarLander, \gls{ape} explanations also suffer from the duplicate issue, where the same condition appears in multiple actions because the Quine-McCluskey algorithm is applied separately to each action. In terms of fidelity metrics other than accuracy in \gls{ret}, \gls{ape} has higher results than \gls{cbs}. Performance-wise, our method has higher total ($E_{CR}$) and average ($E_{AR}$) rewards than \gls{ape} on LunarLander.

\subsection{Varying Parameters}
\label{subsec:varying_parameters}

Changing the predicate definition may generate conflicts with experts' knowledge. For example, an expert defines that a high \gls{sinr} value is above 13 dB, but there are many data points around this value in multiple actions. Thus, shifting the threshold to 10 dB, at which the data points are separable, may lead to different results. 
\Cref{tab:gini_vs_median} compares the evaluation results for the predicate thresholds defined by the Gini(G; \cref{alg:gini_predicate}) and the median(M) for the \gls{cbs} method. 

\begin{table}[h]
    \caption{Evaluation of Median (M) and Gini (G) predicate initiators using \gls{cbs} in different environments.}
    \label{tab:gini_vs_median}
    \centering

\resizebox{\columnwidth}{!}{
\begin{tabular}{|l|r|r|r|r|r|r|r|r|}
\hline
 Env & \multicolumn{2}{|c|}{CartPole} & \multicolumn{2}{|c|}{LunarLander} & \multicolumn{2}{|c|}{MountainCar} & \multicolumn{2}{|c|}{RET} \\
\hline
 Init & G & M & G & M & G & M & G & M \\
\hline
$E_{app}$  & 2.1 & 1.2 & 6.8 & 27.6 & 0.0 & 3.0 & 2.2 & 3.9 \\
$E_{len}$  & 11.0 & 16.1 & 24.4 & 24.5 & 7.9 & 7.3 & 16.0 & 19.7 \\
$E_{dup}$  & 2.7 & 5.0 & 2.9 & 1.8 & 2.7 & 2.4 & 3.2 & 4.1 \\
\hline
$E_{acc}$  & 74.0 & 72.4 & 49.0 & 33.8 & 81.8 & 68.3 & 78.2 & 67.1 \\
$E_{rec}$  & 74.0 & 72.4 & 40.3 & 32.5 & 63.9 & 55.3 & 60.1 & 54.6 \\
$E_{F1}$  & 72.7 & 72.0 & 35.5 & 22.3 & 59.9 & 48.0 & 55.8 & 49.0 \\
\hline
$E_{CR}$  & 14.7 & 20.1 & -35.3 & -35.1 & -76.8 & -89.7 & - & - \\
$E_{TS}$  & 14.7 & 20.1 & 29.2 & 27.3 & 76.8 & 89.7 & - & - \\
$E_{AR}$  & 1.0 & 1.0 & -1.2 & -1.3 & -1.0 & -1.0 & - & - \\
\hline
\end{tabular}
}
\end{table}

The explanations using the Gini-based predicate definition have fewer approximated states ($E_{app}$) than the median, except for CartPole, indicating a better coverage of the explanations. 
In length ($E_{len}$) and duplicated conditions ($E_{dup}$), the results are inconclusive, where there are lower and higher scores in different environments. 
The predicates  defined using Gini purity have higher fidelity scores than the median, showing a better representation of the summarized data. In performance metrics, the proposed Gini-based predicates have higher $E_{CR}$ scores in MountainCar.

\begin{table}[h]
    \caption{Evaluation of different weighting ($w$) functions.} 
    \label{tab:varying_weights}
    \centering
    
\begin{tabular}{|l|r|r|r|r|}
\hline
  $w_{<\text{version}>}$ & 1 & 2 & 3 & 4 \\
\hline
$E_{app}$  & 5.7 & 5.7 & 5.7 & 6.4 \\
$E_{len}$  & 15.9 & 15.9 & 15.9 & 15.9 \\
$E_{dup}$  & 3.1 & 3.1 & 3.1 & 3.1 \\
\hline
$E_{acc}$  & 60.3 & 68.3 & 66.8 & 66.8 \\
$E_{rec}$  & 57.9 & 57.5 & 54.6 & 56.4 \\
$E_{F1}$  & 51.2 & 53.6 & 50.6 & 52.2 \\
\hline
$E_{CR}$  & -37.8 & -31.6 & -32.9 & -32.4 \\
$E_{TS}$  & 46.9 & 43.6 & 40.6 & 40.8 \\
$E_{AR}$  & -0.8 & -0.7 & -0.8 & -0.8 \\
\hline
\end{tabular}
\end{table}

Another parameter that we varied was the weighting function for rule extraction, where we aggregated the scores from all environments because four functions were checked. \cref{tab:varying_weights} shows the results, where all functions have similar $E_{app}$, $E_{len}$, and $E_{dup}$ except that $w_4$ has the highest $E_{app}$. $w_2$ has the highest accuracy, F1, and reward-related ($E_{CR}$ and $E_{AR}$) scores. This is reasonable because $w_2$ captures the percentage of relevant actions under the same condition. Therefore, $w_2$ was selected for subsequent experiments.

\begin{table}[h]
    \caption{Evaluation of different inclusion thresholds ($\theta$).} 
    \label{tab:varying_thresholds}
    \centering
    
\resizebox{\columnwidth}{!}{
\begin{tabular}{|l|r|r|r|r|r|r|r|}
\hline
 $\theta$ & 0.10 & 0.30 & 0.50 & 0.70 & 0.80 & 0.90 & 1.00 \\
\hline
$E_{app}$  & 14.2 & 11.7 & 8.1 & 3.7 & 2.1 & 1.2 & 0.0 \\
$E_{len}$  & 15.9 & 15.9 & 15.9 & 15.9 & 15.9 & 15.9 & 15.9 \\
$E_{dup}$  & 4.4 & 4.0 & 3.5 & 3.1 & 3.0 & 2.3 & 1.5 \\
\hline
$E_{acc}$  & 64.6 & 64.7 & 64.6 & 67.4 & 67.2 & 67.3 & 63.1 \\
$E_{rec}$  & 57.9 & 58.0 & 57.6 & 58.5 & 58.1 & 57.8 & 48.4 \\
$E_{F1}$  & 53.3 & 53.6 & 53.0 & 54.4 & 53.7 & 52.7 & 42.6 \\
\hline
$E_{CR}$  & -29.8 & -32.8 & -34.1 & -33.2 & -34.3 & -34.7 & -37.0 \\
$E_{TS}$  & 42.3 & 43.5 & 43.7 & 42.3 & 41.3 & 41.7 & 46.1 \\
$E_{AR}$  & -0.7 & -0.8 & -0.8 & -0.8 & -0.8 & -0.8 & -0.8 \\
\hline

\end{tabular}
}
\end{table}
The inclusion threshold ($\theta$) is set to determine the number of instances to be included in every cluster that have been ordered by their occurrences. The lower $\theta$, the fewer instances are included to be summarized. \cref{tab:varying_thresholds} presents the results of varying $\theta$ from $10\%$ to full inclusion ($100\%$). Although the length of the explanation ($E_{len}$) is not affected by the threshold, the number of approximated states ($E_{app}$) and duplicated conditions ($E_{dup}$) decrease as $\theta$ increases. This indicates that the coverage of the rules increases as more instances are included. 
In fidelity metrics, the highest scores were obtained at $\theta = 70\%$, which was selected for later experiments in the following subsections. 
A significant drop in the fidelity scores is observed for full inclusion, indicating that infrequent instances are most likely to be outliers. Thus, the strategy of considering frequent instances is justified to achieve higher fidelity scores. 

\subsection{Varying Categories Levels}
\label{subsec:varying_levels}
After varying the parameters above, we selected the following configuration for \gls{cbs} based on the fidelity metrics to continue with the rest of the experiments: Gini-based predicate definition, weighting function $w_2$, and $\theta = 70\%$. In this subsection, we vary the number of levels that a predicate can have from two to seven. The aggregated results for all environments are presented in \cref{tab:varying_levels}. As the number of levels increases, the length of the explanations ($E_{len}$) increases until $n_{cat} = 6$. In contrast, the number of duplicated conditions ($E_{dup}$) decreases, indicating less conflicting information among the different actions. 
The highest fidelity scores were achieved at $n_{cat}$ equal to 5 and 6. 
Although $n_{cat} = 7$ has the highest level-granularity, it has smaller explanations ($E_{len}$) than $n_{cat} = 6$, leading to higher uncovered conditions ($E_{app}$) and lower fidelity scores.

\begin{table}[h]
    \caption{Evaluation of different levels in all environments.}
    \label{tab:varying_levels}
    \centering
    
\resizebox{\columnwidth}{!}{
\begin{tabular}{|l|r|r|r|r|r|r|}
\hline
Levels & 2 & 3 & 4 & 5 & 6 & 7 \\
\hline
$E_{app}$  & 2.1 & 4.0 & 5.6 & 5.0 & 5.5 & 6.7 \\
$E_{len}$  & 14.8 & 15.7 & 17.6 & 20.3 & 21.3 & 20.6 \\
$E_{dup}$  & 2.9 & 1.3 & 1.2 & 0.8 & 0.6 & 0.3 \\
\hline
$E_{acc}$  & 74.7 & 78.5 & 76.8 & 80.2 & 79.8 & 79.4 \\
$E_{rec}$  & 62.0 & 70.3 & 70.3 & 77.6 & 77.7 & 76.4 \\
$E_{F1}$  & 59.1 & 70.4 & 69.9 & 75.1 & 75.1 & 74.3 \\
\hline
$E_{CR}$  & -39.1 & -36.6 & -29.0 & -32.5 & -34.7 & -36.7 \\
$E_{TS}$  & 41.6 & 39.5 & 46.6 & 40.0 & 44.1 & 44.8 \\
$E_{AR}$  & -0.9 & -0.9 & -0.6 & -0.8 & -0.8 & -0.8 \\
\hline
\end{tabular}
}
\end{table}

\begin{table}[h]
    \caption{Evaluation of different levels in MountainCar.}
    \label{tab:varying_levels_mountaincar}
    \centering
    
\resizebox{\columnwidth}{!}{
\begin{tabular}{|l|r|r|r|r|r|r|}
\hline
Levels & 2 & 3 & 4 & 5 & 6 & 7 \\
\hline
$E_{app}$  & 0.0 & 0.0 & 4.5 & 6.3 & 6.3 & 9.0 \\
$E_{len}$  & 7.9 & 12.0 & 13.9 & 15.7 & 21.1 & 19.0 \\
$E_{dup}$  & 2.8 & 3.5 & 3.0 & 1.9 & 2.0 & 1.0 \\
\hline
$E_{acc}$  & 87.9 & 90.0 & 88.2 & 91.6 & 90.5 & 89.0 \\
$E_{rec}$  & 62.7 & 73.7 & 72.7 & 91.6 & 91.1 & 87.8 \\
$E_{F1}$  & 60.9 & 77.5 & 76.4 & 86.9 & 86.2 & 84.5 \\
\hline
$E_{CR}$  & -69.7 & -69.9 & -69.8 & -68.8 & -81.7 & -87.5 \\
$E_{TS}$  & 69.7 & 69.9 & 69.8 & 68.8 & 81.7 & 87.5 \\
$E_{AR}$  & -1.0 & -1.0 & -1.0 & -1.0 & -1.0 & -1.0 \\
\hline
\end{tabular}
}
\end{table}

In performance evaluation, we see higher values of cumulative reward ($E_{CR}$) for $4$, $5$ and $6$ levels. To analyze the impact of more levels/categories, the results for the MountainCar environment are presented in \cref{tab:varying_levels_mountaincar}. 
The cumulative rewards ($E_{CR}$) are significantly higher for $n_{cat} < 6$.
This indicates that the extracted rules could complete the task in certain episodes ($E_{CR}=-100$ indicates failure). With this performance evaluation, the trustworthiness of the explanations is improved because it can be proven that following the explanations can lead to a completed task. Furthermore, if the extracted rules are used to replace the black-box agent, humans can analyze them, and any unwanted condition-action can be corrected or avoided.

\subsection{Varying Refinement Techniques}
In this subsection, we compare the different refinement techniques proposed in \cref{sec:proposed_framework}. Similar to \cref{subsec:varying_levels}, we compared the refinement techniques in all environments and MountainCar, and the results are presented in \cref{tab:varying_refinement}. The baseline for all environments uses the same parameters as in \cref{subsec:varying_levels} with $n_{cat} = 6$ because it achieves the highest F1 and recall, while for MountainCar, we set $n_{cat} = 5$ because it has the highest cumulative rewards. We aim to analyze whether the proposed refinement techniques can further improve the results. We used a maximum recommendation ($m$) of 5, an iteration budget ($b$) of 5 to minimize duplicates, and 10 to maximize F1 with an adjustment rate ($\alpha$) of 0.5.

\begin{table}[h]
    \caption{Evaluation of different refinement techniques for two different setups.}
    \label{tab:varying_refinement}
    \centering
    
\resizebox{\columnwidth}{!}{
\begin{tabular}{|c|r|r|r|r|r|r|}
\hline
Levels & \multicolumn{3}{|c|}{All Environments - 6 levels} & \multicolumn{3}{|c|}{MountainCar - 5 levels} \\
\hline
Refine  & base & $min_{du}$ & $max_{F1}$ & base & $min_{du}$ & $max_{F1}$ \\
\hline
$E_{app}$  & 5.46 & 5.22 & 4.50 & 6.34 & 3.63 & 1.99 \\
$E_{len}$  & 21.32 & 20.62 & 22.30 & 15.70 & 15.60 & 18.70 \\
$E_{dup}$  & 0.60 & 0.35 & 0.80 & 1.90 & 0.90 & 3.50 \\
\hline
$E_{acc}$  & 79.75 & 79.90 & 80.88 & 91.56 & 92.76 & 93.66 \\
$E_{rec}$  & 77.67 & 77.15 & 78.51 & 91.63 & 91.36 & 92.41 \\
$E_{F1}$  & 75.09 & 75.09 & 76.18 & 86.95 & 87.68 & 88.90 \\
\hline
$E_{CR}$  & -31.70 & -35.56 & -30.69 & -68.75 & -63.20 & -63.40 \\
$E_{TS}$  & 57.86 & 58.34 & 54.44 & 68.75 & 63.20 & 63.40 \\
$E_{AR}$  & -0.55 & -0.61 & -0.56 & -1.00 & -1.00 & -1.00 \\
\hline
\end{tabular}
}
\end{table}

In both setups, the minimize-duplicate technique ($min_{du}$) successfully reduced the duplicated conditions ($E_{dup}$). It also reduced the length of the explanation ($E_{len}$) and the approximated states ($E_{app}$). This refinement technique also improved $E_{acc}$ in both setups. Although $E_{CR}$ in MountainCar is improved, the aggregated results from all environments have lower scores.

On the other hand, the maximize-F1 technique ($max_{F1}$) increases the length of the explanation ($E_{len}$) in both setups. As reducing $E_{dup}$ is not the focus of this technique, this score increased in both setups. The goal of improving the F1 score was achieved, as shown by the improvement in all fidelity scores. Interestingly, cumulative reward scores ($E_{CR}$) are also improved, and in MountainCar the score could be further improved by more than five points.

\subsection{\gls{ret} Results}
\label{subsec:ret_results}

For the \gls{ret} environment, we selected a few configurations, as shown in \cref{tab:ret_performance}, along with the black-box agent (DQNRD). With six levels of categories, the explanations from \gls{cbs} achieve a higher fidelity than those from \gls{ape}. 
To obtain a representative explanation of the \gls{ret} agent, six levels of categories should be used because they have high fidelity scores. 
Performance-wise, they are all relatively similar and show slightly compromised performance ($E_{AR}$), considering that the maximum value is 1 ($10$ when multiplied). 
If we want to replace the black-box agent with a transparent one, \gls{cbs} with binary categories has the least compromised performance.
Furthermore, by showing the reward components, we observe that \gls{rsrp} is improved in this setup. In this way, the operator can anticipate the cost/gain of different factors when replacing the black-box agent with a transparent one. 

\begin{table}[h]
    \caption{Evaluation of the RET environment.} 
    \label{tab:ret_performance}
    \centering

\resizebox{\columnwidth}{!}{
\begin{tabular}{|l|r|r|r|r|r|}
\hline
  Method & APE & \multicolumn{3}{|c|}{CBS} & DQNRD \\
% \hline
%  Levels & 2 & 2 & \multicolumn{2}{|c|}{6} & - \\
% \hline
%  Initiator & M & M & \multicolumn{2}{|c|}{G} & - \\
\hline
 Initiator; Levels & M;2 & M;2 & \multicolumn{2}{|c|}{G;6} & - \\
\hline
 $ \theta$ & 1.0 & 0.7 & \multicolumn{2}{|c|}{0.7} & - \\
\hline
 Refinement & none & $min_{du}$ & $max_{F1}$ & none & - \\
\hline
$E_{app}$  & 0.00 & 2.30 & 2.86 & 2.95 & - \\
$E_{len}$  & 17.00 & 16.80 & 20.00 & 21.40 & - \\
$E_{dup}$  & 0.00 & 2.30 & 0.00 & 0.00 & - \\
\hline
$E_{acc}$  & 72.02 & 70.43 & 80.17 & 78.86 & - \\
$E_{rec}$  & 62.41 & 63.40 & 72.93 & 72.15 & - \\
$E_{F1}$  & 60.17 & 59.34 & 70.74 & 69.50 & - \\
\hline
$E_{CR}$ & 10.54 & 10.88 & 10.10 & 9.94 & 11.08 \\
$E_{TS}$  & 20 & 20 & 20 & 20 & 20 \\
$E_{AR}$ ($	\times 10$) & 5.27 & 5.44 & 5.05 & 4.97 & 5.54 \\
$E_{AR}$ (RSRP $\times 10$) & 2.85 & 2.93 & 2.82 & 2.79 & 2.89 \\
$E_{AR}$ (SINR $\times 10$) & 2.42 & 2.51 & 2.23 & 2.18 & 2.65 \\
\hline
\end{tabular}
}
\end{table}

\section{Discussion}
\label{sec:discussion}

% LLM part
In this work, we demonstrated that \glspl{llm} are suitable for interpreting questions to trigger summarization. We did not use an \gls{llm} to perform data summarization because \glspl{llm} are designed to process textual information rather than numerical data. 
In the agentic \gls{ai} era, the proposed framework has the potential to become a building block for explainer agents. 
This saves resources by not feeding lengthy numerical data to a highly complex \gls{llm}, thereby avoiding its computationally intensive inference step.

When all conditions appear in one queried action, \gls{ape} cannot generate any informative explanation. Our \gls{cbs} method addresses this by focusing on frequent states to always generate explanations successfully. 
We also showed that categorical predicates (not binary) improved the quality of explanations. 
\gls{ape} utilizes Quine-McCluskey with a computational complexity of $\frac{3^{|P|}}{|P|}$ that grows with the number of predicates ($|P|$). On the other hand, \gls{cbs} uses the clustering method with complexity $O(|S| \times |P| \times k^2 \times i)$, where $|S|$ is the size of the data, $i$ is the iteration, and $k$ is the size of the cluster.
Thus, \gls{ape} can grow exponentially with respect to the discretization granularity, whereas \gls{cbs} grows linearly. 
Furthermore, our method can compress the explanation by combining multiple predicate values into a single statement.

Our proposed explanation-to-rules conversion can be used to evaluate the correctness and potentially replace the black-box agent when performance trade-offs are acceptable.
By deploying the rules in the environment, they are evaluated to determine whether performance remains high or is compromised. 
Due to the Markovian nature of \gls{rl}'s formulation, a few mistakes can significantly affect the final outcome. This occurred for the rules in the CartPole and LunarLander environments, where the correct actions should have been taken in certain critical states. Additionally, the explanations/rules imitated the historical data (not directly learned in the environment) with limited expressivity. 
On the other hand, the extracted rules in the MountainCar environment could complete the task. It shows that the extracted rules can replace the black-box agent to perform the task.
Similarly, in the \gls{ret} environment, replacing the black-box agent with the extracted rules slightly compromised performance. 
Furthermore, the evaluation scores in these two environments can be improved using the proposed refinement techniques. 
This shows that transparent rules with satisfactory performance can be extracted using \gls{cbs}, whereas a more complex model (e.g., deep \gls{rl}) is still required for more complex tasks.

The proposed rule extraction method enables fidelity evaluation of textual explanations. 
These metrics are important for evaluating the explanation quality for retrospective analysis because they do not affect the decision-making process. We have shown through the experiments how the parameters affected and improved the fidelity scores. The improvement in the fidelity score is crucial for understanding complex model behavior, as some of them cannot be replaced by transparent ones. Furthermore, the coherence of the explanations was indicated by duplicate conditions, which were minimized through a refinement technique.

\section{Conclusion and Future Work}
\label{sec:conclusion}

In this article, we have presented a framework for explaining a trained \gls{rl} policy using historical data by generating textual explanations, extracting rules, and evaluating their properties, fidelity, and performance. This framework can support real-time operator/expert decision-making (e.g., retrospective analysis of an agent’s tilt recommendation before execution). 
Their knowledge can be incorporated into predicate definitions, and automatic generation has also been proposed as an alternative.  
The evaluation procedure provides systematic measures of textual \gls{rl} explanation quality. 
We have shown that the proposed \gls{cbs} addresses the issues in \gls{ape} with comparable performance. 
Predicate generation based on the Gini value has better fidelity than statistical quantiles. In addition, the proposed refinement techniques could further improve the explanation quality. 
With the proposed method, the black-box model can be replaced with a transparent one if the performance loss is tolerable.

% Future work 
This work can be extended to find anomalies by inverting the rules in the clustering. 
The framework can also be used for supervised learning problems, where performance evaluation is no longer relevant, but fidelity evaluation can be performed directly with ground truth labels. Although extending this work with attributive explanations is direct, integrating other types of explanation, such as contrastive or counterfactual, requires further investigation. Alternatively, a potential extension is to incorporate non-tabular states, such as visual inputs, via concept-embedding techniques. Another possible future direction is to investigate the adaptability of the transparent agent to continue learning, resembling the \gls{rl} agent on deployment.

\section*{Acknowledgement}
This work was partially supported by the Wallenberg AI, Autonomous Systems and Software Program (WASP) funded by the Knut and Alice Wallenberg Foundation and the TECoSA Centre for Trustworthy Edge Computing Systems and Applications. The authors thank Agustín Valencia and Alexandros Nikou for providing valuable feedback that helped improve the clarity of the paper.

\bibliographystyle{IEEEtran}
\bibliography{setup/bibliography}

\end{document}